\documentclass[letterpaper, 10 pt, conference]{ieeeconf}
\IEEEoverridecommandlockouts

\usepackage[colorlinks]{hyperref}
\hypersetup{
    colorlinks,
    linkcolor=black,
    citecolor=black,
    filecolor=magenta,
    urlcolor=cyan,
}

\usepackage{resizegather}

\usepackage{cite}
\usepackage{amsmath,amssymb,amsfonts,mathrsfs}
\usepackage{algorithmic}
\usepackage{graphicx}
\usepackage{textcomp}
\usepackage{xcolor}
\usepackage{tcolorbox}

\usepackage{tablefootnote}
\usepackage{threeparttable}
\usepackage{caption, subcaption}

\usepackage{tikz}
\usepackage{graphicx}
\usepackage{tkz-euclide}

\usetikzlibrary{positioning}
\usetikzlibrary{shapes}
\usetikzlibrary{shapes.misc}
\usetikzlibrary{shapes.geometric}
\usetikzlibrary{plotmarks}
\usetikzlibrary{intersections}
\usetikzlibrary{calc}
\usetikzlibrary{fit}
\usetikzlibrary{patterns,tikzmark}
\usetikzlibrary{matrix,decorations.pathreplacing,calc}

\tikzset{cross/.style={cross out, draw, 
         minimum size=2*(#1-\pgflinewidth), 
         inner sep=0pt, outer sep=0pt}}

\newcommand{\state}[0]{x}
\newcommand{\prel}[0]{p_{\rm{rel}}}
\newcommand{\vrel}[0]{v_{\rm{rel}}}
\newcommand{\preldot}[0]{\dot{p}_{\rm{rel}}}

\newcommand{\norm}[1]{\left\lVert#1\right\rVert}

\newcommand{\vertiii}[1]{{\left\vert\kern-0.25ex\left\vert\kern-0.25ex\left\vert #1 
    \right\vert\kern-0.25ex\right\vert\kern-0.25ex\right\vert}}

\newtheorem{theorem}{Theorem}

\newtheorem{remark}{Remark}

\newtheorem{definition}{Definition}

\makeatletter
\renewcommand{\fps@figure}{htp}
\renewcommand{\fps@table}{htp}
\makeatother

\def\BibTeX{{\rm B\kern-.05em{\sc i\kern-.025em b}\kern-.08em
    T\kern-.1667em\lower.7ex\hbox{E}\kern-.125emX}}
    
\begin{document}

\title{Collision Cone Control Barrier Functions: Experimental Validation on UGVs for Kinematic Obstacle Avoidance}

\author{Bhavya Giri Goswami$^{*1}$, Manan Tayal$^{*1}$, Karthik Rajgopal$^{1}$, Pushpak Jagtap$^{1}$, Shishir Kolathaya$^{1}$
\thanks{This research was supported by the Wipro IISc Research Innovation Network (WIRIN).
}
\thanks{$^*$These authors have contributed equally.
}
\thanks{$^{1}$Robert Bosch Center for Cyber-Physical Systems (RBCCPS), Indian Institute of Science (IISc), Bengaluru.
{\tt\scriptsize bhavya.goswami@uwaterloo.ca \{manantayal, karthikrajgo, pushpak, shishirk\}@iisc.ac.in} 
}%
}

\maketitle
\begin{abstract}
Autonomy advances have enabled robots in diverse environments and close human interaction, necessitating controllers with formal safety guarantees. This paper introduces an experimental platform designed for the validation and demonstration of a novel class of Control Barrier Functions (CBFs) tailored for Unmanned Ground Vehicles (UGVs) to proactively prevent collisions with kinematic obstacles by integrating the concept of collision cones. While existing CBF formulations excel with static obstacles, extensions to torque/acceleration-controlled unicycle and bicycle models have seen limited success. Conventional CBF applications in nonholonomic UGV models have demonstrated control conservatism, particularly in scenarios where steering/thrust control was deemed infeasible. Drawing inspiration from collision cones in path planning, we present a pioneering CBF formulation ensuring theoretical safety guarantees for both unicycle and bicycle models. The core premise revolves around aligning the obstacle's velocity away from the vehicle, establishing a constraint to perpetually avoid vectors directed towards it. This control methodology is rigorously validated through simulations and experimental verification on the Copernicus mobile robot (Unicycle Model) and FOCAS-Car (Bicycle Model).

\end{abstract}


\section{Introduction}
\label{section: Introduction}


\par The progress in autonomous technologies has facilitated the deployment of robots in a wide range of environments, often in close proximity to humans. Consequently, the design of controllers with formally assured safety has become a crucial aspect of these safety-critical applications, constituting an active area of research in recent times. Researchers have devised various methodologies to address this challenge, including model predictive control (MPC) \cite{yu_mpc_aut_ground_vehicle}, reference governor \cite{6859176}, reachability analysis \cite{8263977} \cite{https://doi.org/10.48550/arxiv.2106.13176}, and artificial potential fields \cite{Singletary2021ComparativeAO}. To establish formal safety guarantees, such as collision avoidance with obstacles, it is essential to employ a safety-critical control algorithm that encompasses both trajectory tracking/planning and prioritizes safety over tracking. One such approach is based on control barrier functions [6] (CBFs), which define a secure state set through inequality constraints and formulate it as a quadratic programming (QP) problem to ensure the forward invariance of these sets over time.

A significant advantage of using CBF-based quadratic programs over other state-of-the-art techniques is that they work efficiently on real-time practical applications in complex dynamic environments \cite{Singletary2021ComparativeAO, https://doi.org/10.48550/arxiv.2106.13176}; that is, optimal control inputs can be computed at a very high frequency on off-the-shelf electronics. It can be applied as a fast safety filter over existing path planning controllers\cite{9682803}.
They are already being used in UGVs for collision avoidance as shown in \cite{7040372, https://doi.org/10.48550/arxiv.2103.12382, 7864310, 9029446, 9112342, 9565037}. Many contributions shown here are for point mass dynamic models with static obstacles, and their extensions for unicycle models are shown in \cite{XIAO2021109592, https://doi.org/10.48550/arxiv.2106.05341, 9482979}. However, these extensions are shown for velocity-controlled models, not for acceleration-controlled models. 
On the other hand, the Higher Order CBF (HOCBF) based approaches mentioned in \cite{9516971} are shown to successfully avoid collisions with static obstacles using acceleration-controlled unicycle models but lack geometrical intuition.
Extension of this framework (HOCBF) for the case of moving obstacles is possible, however, safety guarantees are provided for a subset of the original safe set, thereby making it conservative. In addition, computation for a feasible CBF candidate with apt penalty and parameter terms is not that straightforward (as discussed at the end of Section-\ref{section: Background}).

Thus, two major challenges remain that prevent the successful deployment of these CBF-QPs for obstacle avoidance in a dynamic environment: a) Existing CBFs are not able to handle the nonholonomic nature of UGVs well. They provide limited control capability in the acceleration-controlled unicycle and bicycle models, i.e., the solutions from the CBF-QPs have either no steering or forward thrust capabilities (this is shown in Section \ref{section: Background}), and b) Existing CBFs are not able to handle dynamic obstacles well, i.e., the controllers are not able to avoid collisions with moving obstacles, which is also shown in Section \ref{section: Background}.

%

\begin{figure}[t]
    \centering
    \includegraphics[width=0.44\linewidth]{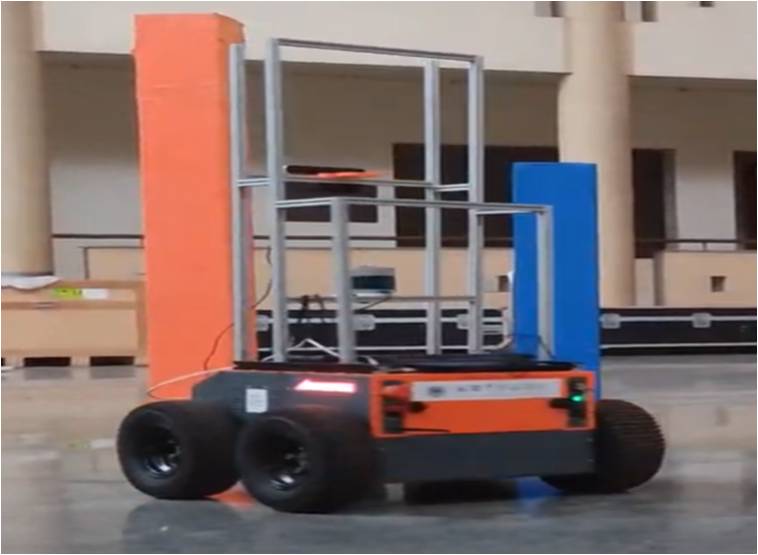}
    \includegraphics[width=0.46\linewidth]{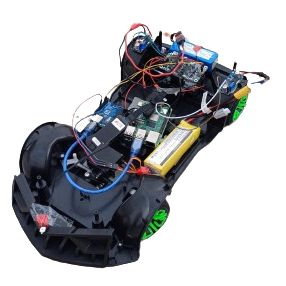}
\caption{Test setups: Copernicus (left); FOCAS-Car (right). }
\label{fig:Copernicus and FOCAS-Car}
\end{figure}

Intending to address the above challenges, we propose a new class of CBFs via the concept of collision cones. In particular, we generate a new class of constraints that ensure that the relative velocity between the obstacle and the vehicle always points away from the direction of the vehicle's approach. Assuming ellipsoidal shape for the obstacles \cite{709600}, the resulting set of unwanted directions for potential collision forms a conical shape, giving rise to the synthesis of \textbf{Collision Cone Control Barrier Functions (C3BFs)}. The C3BF-based QP optimally and rapidly calculates inputs in real-time such that the relative velocity vector direction is kept out of the collision cone for all time. This approach is demonstrated using the acceleration-controlled non-holonomic models (Fig. \ref{fig:Copernicus and FOCAS-Car}). 
\label{section: Literature Review}


The idea of collision cones was first introduced in \cite{Fiorini1993, doi:10.1177/027836499801700706, 709600} as a means to geometrically represent the possible set of velocity vectors of the vehicle that lead to a collision. 
The approach was extended for irregularly shaped robots and obstacles with unknown trajectories both in 2D \cite{709600} and 3D space \cite{Chakravarthy2012}. 
This was commonly used for offline obstacle-free path planning applications \cite{doi:10.2514/1.G005879, doi:10.2514/6.2004-4879} like missile guidance. Collision cones have already been incorporated into MPC by defining the cones as constraints \cite{babu2018model}, but making it a CBF-based constraint has yet to be addressed. 

\subsection{Contribution and Paper Structure}
The main contribution of our work is to include the velocity information of the obstacles in the CBFs, thereby yielding a 
real-time dynamic obstacle avoidance controller.
Our specific contributions include the following:

\begin{itemize}
    \item We formulate a new class of CBFs by using the concept of collision cones. The resulting CBF-QP can be computed in real-time, thereby yielding a safeguarding controller that avoids collision with moving obstacles. This can be stacked over any state-of-the-art planning algorithm with the C3BF-QP acting as a safety filter.
    \item We formally show how the proposed formulation yields a safeguarding control law for wheeled robots with nonholonomic constraints (i.e. acceleration-controlled unicycle \& bicycle models). Note that much of the existing works on CBFs for mobile robots are with point mass models, and our contributions here are for nonholonomic wheeled robots, which are more challenging to control. We demonstrate our solutions in both simulations and hardware experiments.
\end{itemize}




\section{Background}
\label{section: Background}
In this section, we provide the relevant background necessary to formulate our problem of moving obstacle avoidance. Specifically, we first describe the vehicle models considered in our work, namely the acceleration-controlled unicycle and the bicycle models. Next, we formally introduce Control Barrier Functions (CBFs) and their importance for real-time safety-critical control for considered vehicle models. Finally, we explain the shortcomings in existing CBF approaches in the context of collision avoidance of moving obstacles.

\begin{figure}[htbp]
    \centering
    \includegraphics[width=0.4\linewidth]{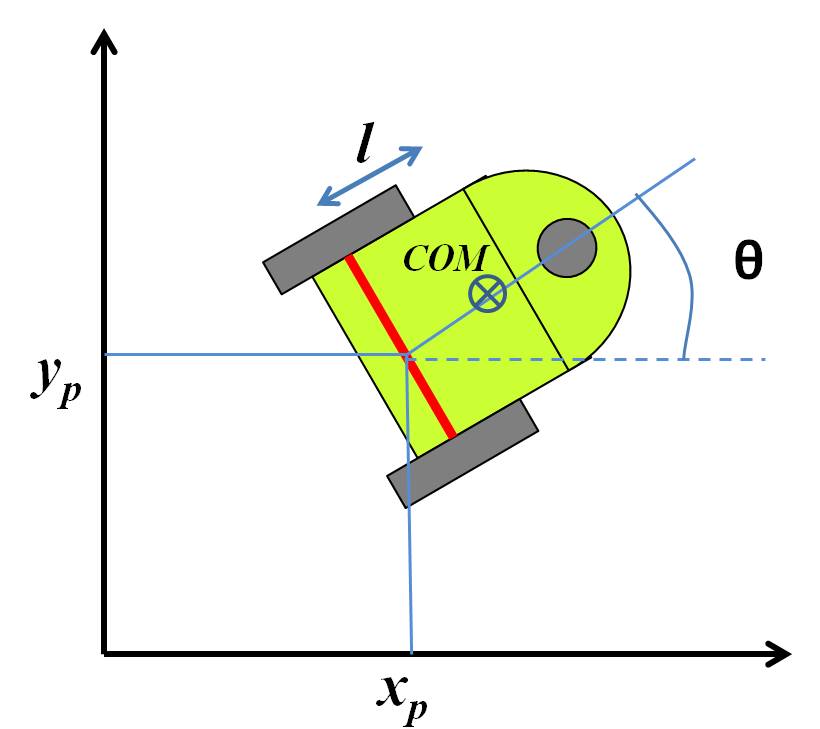}
    \includegraphics[width=0.52\linewidth]{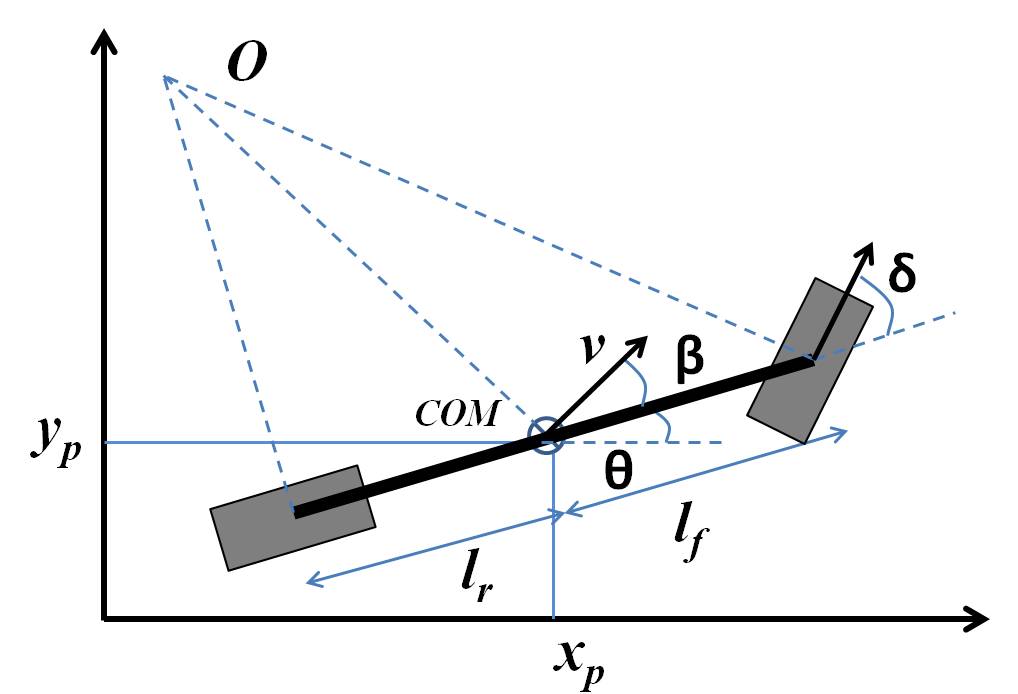}
\caption{Schematic of Unicycle (left); Bicycle model (right).}
\label{fig:models}
\end{figure}

\subsection{Vehicle Models}

\subsubsection{Acceleration controlled unicycle model}
\par The unicycle model encompasses state variables denoted as $x_p$, $y_p$, $\theta$, $v$, and $\omega$, representing pose, linear velocity, and angular velocity, respectively. Linear acceleration $(a)$ and angular acceleration $(\alpha)$ serve as the control inputs. In Figure \ref{fig:models}, a differential drive robot is depicted, which is effectively described by the unicycle model presented below:
\begin{equation}
	\begin{bmatrix}
		\dot{x}_p \\
		\dot{y}_p \\
		\dot{\theta} \\
		\dot{v} \\
		\dot{\omega}
	\end{bmatrix}
	=
        \begin{bmatrix}
            v\cos\theta\\
            v\sin\theta\\
            \omega \\
            0 \\
            0
        \end{bmatrix}
	+
	\begin{bmatrix}
            0 & 0 \\
            0 & 0 \\
            0 & 0 \\
            1 & 0 \\
            0 & 1
	\end{bmatrix}
	\begin{bmatrix}
		a \\
		\alpha
	\end{bmatrix}.
    \label{eqn:Acceleration controlled Unicycle model}
\end{equation}

While the conventional unicycle model found in the literature typically employs linear and angular velocities, denoted as $v$ and $\omega$, as inputs, we opt for accelerations as inputs due to the rationale outlined in \cite{thontepu2023control}.

\subsubsection{Acceleration controlled bicycle model}

The bicycle model comprises two wheels, with the front wheel dedicated to steering (refer to Fig. \ref{fig:models}). This model finds common application in the domain of self-driving cars, where we simplify the treatment of front and rear wheel sets as a unified virtual wheel for each set.
The dynamics of the bicycle model are outlined as follows:
\begin{align}
    \begin{bmatrix}
        \dot x_p \\
        \dot y_p \\
        \dot \theta \\
        \dot v 
    \end{bmatrix} 
    & = 
    \begin{bmatrix}
        v \cos (\theta + \beta) \\
        v \sin (\theta + \beta) \\
        \frac{v}{l_r} \sin (\beta) \\
        a 
    \end{bmatrix},
    \label{eq:bicyclemodel} \\
	\text{where} \quad \beta &= \tan^{-1}\left(\frac{l_r}{l_f + l_r}\tan(\delta)\right), \label{eq:SlipSteeringConv}
\end{align}
$x_p$ and $y_p$ denote the coordinates of the vehicle’s center of mass (CoM) in an inertial frame. $\theta$ represents the orientation of the vehicle with respect to the $x$ axis. $a$ is the linear acceleration at CoM.
$l_f$ and $l_r$ are the distances of the front and rear axles from the CoM,  respectively.
$\delta$ is the steering angle of the vehicle and 
$\beta$ is the vehicle's slip angle, i.e., the steering angle of the vehicle mapped to its CoM (see Fig. \ref{fig:models}).
This is not to be confused with the tire slip angle. 

\begin{remark}
Similar to \cite{https://doi.org/10.48550/arxiv.2103.12382}, we assume that the slip angle is constrained to be small. As a result, we approximate $\cos \beta \approx 1$ and $\sin \beta \approx \beta$. Accordingly, we get the following simplified dynamics of the bicycle model:
\begin{equation}
\label{eqn:bicyclemodel with small beta}
	\underbrace{\begin{bmatrix}
		\dot{x}_p \\
		\dot{y}_p \\
		\dot{\theta} \\
		\dot{v}
	\end{bmatrix}}_{\dot{\state}}
	=
	\underbrace{\begin{bmatrix}
		v \cos\theta \\
		v \sin \theta \\
		0 \\
		0
	\end{bmatrix}}_{f(\state)}
	+
	\underbrace{\begin{bmatrix}
		0 & - v\sin\theta \\
		0 & v\cos\theta \\
		0 & \frac{v}{l_r} \\
		1 & 0
	\end{bmatrix}}_{g(\state)}
	\underbrace{\begin{bmatrix}
		a \\
		\beta
	\end{bmatrix}}_{u}.
\end{equation}
Since the control inputs $a, \beta$ are now affine in the dynamics, CBF-QPs can be constructed directly to yield real-time control laws, as explained next.
\end{remark}


\subsection{Control barrier functions (CBFs)}

After providing an overview of the vehicle models, we proceed to formally introduce control barrier functions and their application in ensuring safety within the context of a non-linear control system described by the affine equation:
\begin{equation}
	\dot{\state} = f(\state) + g(\state)u.
	\label{eqn: affine control system}
\end{equation}
Here, \(\state\) belongs to the set \(\mathcal{D} \subseteq \mathbb{R}^n\) representing the system's state, and \(u\) lies in \(\mathbb{U} \subseteq \mathbb{R}^m\) representing the system's input. We assume that the functions \(f: \mathbb{R}^n \rightarrow \mathbb{R}^n\) and \(g: \mathbb{R}^n \rightarrow \mathbb{R}^{n \times m}\) are locally Lipschitz. Given a Lipschitz continuous control law \(u = k(\state)\), the resulting closed-loop system \(\dot{\state} = f_{cl}(\state) = f(\state) + g(\state)k(\state)\) leads to a solution \(\state(t)\), with an initial condition of \(\state(0) = \state_0\).

Consider a set $\mathcal{C}$ defined as the \textit{super-level set} of a continuously differentiable function $h:\mathcal{D}\rightarrow \mathbb{R}$ yielding,
\begin{align}
\label{eq:setc1}
	\mathcal{C}                        & = \{ \state \in \mathcal{D} \subset \mathbb{R}^n : h(\state) \geq 0\} \\
\label{eq:setc2}
	\partial\mathcal{C}                & = \{ \state \in \mathcal{D} \subset \mathbb{R}^n : h(\state) = 0\}\\
\label{eq:setc3}
	\text{Int}\left(\mathcal{C}\right) & = \{ \state \in \mathcal{D} \subset \mathbb{R}^n : h(\state) > 0\}.
\end{align}

It is assumed that the interior of \(\mathcal{C}\) is non-empty, and \(\mathcal{C}\) has no isolated points, i.e., \(\text{Int}(\mathcal{C}) \neq \emptyset\) and the closure of the interior of \(\mathcal{C}\) is equal to \(\mathcal{C}\). The system is deemed safe with respect to the control law \(u = k(\state)\) if for any \(\state(0) \in \mathcal{C}\), it follows that \(\state(t) \in \mathcal{C}\) for all \(t \geq 0\).

We can ascertain the safety provided by the controller \(k(\state)\) through the use of control barrier functions (CBFs), as defined below.
\begin{definition}[Control barrier function (CBF)]{\it
\label{definition: CBF definition}
Given the set $\mathcal{C}$ defined by \eqref{eq:setc1}-\eqref{eq:setc3}, with $\frac{\partial h}{\partial \state}(\state) \neq 0\; \forall \state \in \partial \mathcal{C}$, the function $h$ is called the control barrier function (CBF) defined on the set $\mathcal{D}$, if there exists an extended \textit{class} $\mathcal{K}$ function\footnote{A continuous function $\kappa : [0, d) \rightarrow [0, \infty)$ for some $d > 0$ is said to belong to class-$\mathcal{K}$ if it is strictly increasing and $\kappa(0) = 0$. Here, $d$ is allowed to be $\infty$. The same function can be extended to the interval $\kappa: (-b,d)\to (-\infty, \infty)$ with $b>0$ (which is also allowed to be $\infty$), in which case we call it the extended class $\mathcal{K}$ function.} $\kappa$ such that for all $\state \in \mathcal{D}$:
\begin{equation}
\begin{aligned}
    \sup_{ u \in \mathbb{U}}\! \left[\underbrace{\mathcal{L}_{f} h(\state) + \mathcal{L}_g h(\state)u} \iffalse+ \frac{\partial h}{\partial t}\fi_{\dot{h}\left(\state, u\right)} \! + \kappa\left(h(\state)\right)\right] \! \geq \! 0
\end{aligned}
\end{equation}
where $\mathcal{L}_{f} h(\state) := \frac{\partial h}{\partial \state}f(\state)$ and $\mathcal{L}_{g} h(\state):= \frac{\partial h}{\partial \state}g(\state)$ are the Lie derivatives. 

}
\end{definition}
As per \cite{Ames_2017} and \cite{8796030}, any Lipschitz continuous control law \(k(\state)\) that satisfies the inequality \(\dot{h} + \kappa( h )\geq 0\) ensures the safety of \(\mathcal{C}\) given \(\state(0)\in \mathcal{C}\), and it leads to asymptotic convergence towards \(\mathcal{C}\) if \(\state(0)\) is outside of \(\mathcal{C}\). Notably, CBFs can also be defined solely on \(\mathcal{C}\), ensuring only safety. This will find utility in the bicycle model \eqref{eqn:bicyclemodel with small beta}, which is elaborated upon later.

\subsection{Controller synthesis for real-time safety}
Having described the CBF and its associated formal results, we now discuss its Quadratic Programming (QP) formulation. 
CBFs are typically regarded as \textit{safety filters} which take the desired input (reference controller input) $u_{ref}(\state,t)$ and modify this input in a minimal way: 

\begin{equation}
\begin{aligned}
\label{eqn: CBF QP}
u^{*}(x,t) &= \min_{u \in \mathbb{U} \subseteq \mathbb{R}^m} \norm{u - u_{ref}(x,t)}^2\\
\quad & \textrm{s.t. } \mathcal{L}_f h(x) + \mathcal{L}_g h(x)u + \kappa \left(h(x)\right) \geq 0\\
\end{aligned}
\end{equation}

This is called the Control Barrier Function based Quadratic Program (CBF-QP). The reference control input can be from any SOTA algorithm. If $\mathbb{U}=\mathbb{R}^m$, then the QP is feasible, and the explicit solution is given by
\begin{equation*}
	u^{*}(x, t) = u_{ref}(x, t) + u_{safe}(x,t),
\end{equation*}
where $u_{safe}(x,t)$ is given by
\begin{multline}\label{eq:CBF-QP}
u_{safe}(x, t) \!=\!
	\begin{cases}
		0 & \text{for } \psi(x, t) \geq 0 \\
		-\frac{\mathcal{L}_{g}h(x)^T \psi(x, t)}{\mathcal{L}_{g}h(x)\mathcal{L}_{g}h(x)^T} & \text{for } \psi(x, t) < 0,
	\end{cases}
\end{multline}
where $\psi (x,t) := \dot{h}\left(x, u_{ref}(x, t)\right) + \kappa \left(h(x)\right)$. The sign change of $\psi$ yields a switching type of control law.

\subsection{Classical CBFs and moving obstacle avoidance}
Having introduced CBFs, we now explore collision avoidance in Unmanned Ground Vehicles (UGVs). In particular, we discuss the problems associated with the classical CBF-QPs, especially with the velocity obstacles. 
We also summarize and compare with C3BFs in Table \ref{table: types of CBFs}.

\begin{table*}[t]
\caption{Comparison between the Ellipse CBF \eqref{eqn:Ellipse-CBF}, HOCBF \eqref{eq:hocbf} and the proposed C3BF \eqref{eqn:CC-CBF} for different UGV models.}
\begin{center}
    \begin{threeparttable}
	\begin{tabular}{| c | c | c | c |}
		\hline
		\textbf{CBFs} & \textbf{Vehicle Models} & \textbf{Static Obstacle} $(c_x, c_y)$ & \textbf{Moving Obstacle} $(c_x(t), c_y(t)) ^\dag$ \\
		\hline
		\textbf{Ellipse CBF}  & Acceleration-controlled unicycle \eqref{eqn:Acceleration controlled Unicycle model}   & Not a valid CBF                        & Not a valid CBF                               \\
		\hline
		\textbf{Ellipse CBF}  & Bicycle \eqref{eqn:bicyclemodel with small beta}    & Valid CBF, No acceleration                        & Not a valid CBF                               \\
		\hline
		\textbf{HOCBF}      & Acceleration-controlled unicycle  \eqref{eqn:Acceleration controlled Unicycle model}  & Valid CBF, No steering                        & Valid CBF, but conservative  \\
 		\hline
   		\textbf{HOCBF}      & Bicycle \eqref{eqn:bicyclemodel with small beta}  & Valid CBF                        & Not a valid CBF  \\
 		\hline
		\textbf{C3BF}   & Acceleration-controlled unicycle \eqref{eqn:Acceleration controlled Unicycle model} & Valid CBF in $\mathcal{D}$                       & Valid CBF in $\mathcal{D}$                              \\
		\hline
            \textbf{C3BF} & Bicycle \eqref{eqn:bicyclemodel with small beta} & Valid CBF in $\mathcal{C}$ & Valid CBF in $\mathcal{C}$ \\
            \hline
	\end{tabular}
	\begin{tablenotes}\footnotesize
    \item[$\dag$] $(c_x(t), c_y(t))$ are continuous (or at least piece-wise continuous) functions of time
    \end{tablenotes}
	\end{threeparttable}
\end{center}
\label{table: types of CBFs}
\end{table*}

\subsubsection{Ellipse-CBF Candidate - Unicycle}
Consider the following CBF candidate:
\begin{equation}
    h(\state,t) = \left(\frac{c_x(t) - x_p}{c_1}\right)^2 + \left(\frac{c_y(t) - y_p}{c_2}\right)^2 - 1,
    \label{eqn:Ellipse-CBF}
\end{equation}
which approximates an obstacle with an ellipse with center $(c_x(t), c_y(t))$ and axis lengths $c_1,c_2$. We assume that $c_x(t),c_y(t)$ are differentiable and their derivatives are piece-wise constants. 
Since $h$ in \eqref{eqn:Ellipse-CBF} is dependent on time (e.g., moving obstacles), the resulting set $\mathcal{C}$ is also dependent on time. To analyze this class of sets, time-dependent versions of CBFs can be used \cite{IGARASHI2019735}. Alternatively, we can reformulate our problem to treat the obstacle position $c_x,c_y$ as states, with their derivatives being constants. This will allow us to continue using the classical CBF given by Definition \ref{definition: CBF definition}, including its properties on safety. The derivative of \eqref{eqn:Ellipse-CBF} is
\begin{align}
\dot h=\frac{2(c_x-x_p)(\dot c_x - v\cos \theta)}{c_1^2} +\frac{2(c_y-y_p)(\dot c_y - v \sin \theta)}{c_2^2},
\end{align}
which has no dependency on the inputs $a, \alpha$. Hence, $h$ will not be a valid CBF for the acceleration-based model \eqref{eqn:Acceleration controlled Unicycle model}. 

However, for static obstacles, if we choose to use the velocity-controlled model (with $v,\omega$ as inputs instead of $a,\alpha$), then $h$ will certainly be a valid CBF, but the vehicle will have limited control capability, i.e., it looses steering $\omega$.

\subsubsection{Ellipse-CBF Candidate - Bicycle}
For the bicycle model, the derivative of $h$ \eqref{eqn:Ellipse-CBF} yields
\begin{align}
    \dot h = & 2 (c_x - x_p) ( \dot c_x - v \cos \theta + v (\sin \theta) \beta)/c_1^2 \nonumber \\
    & + 2 (c_y - y_p) ( \dot c_y - v \sin \theta - v (\cos \theta) \beta)/c_2^2,
\end{align}
which only has $\beta$ as the input. Furthermore, the derivatives $\dot c_x, \dot c_y$ are free variables, i.e., the obstacle velocities can be selected in such a way that the constraint $\dot h(x,u)+ \kappa (h(x)) < 0$, whenever $\mathcal{L}_g h=0$. This implies that $h$ is not a valid CBF for moving obstacles for the bicycle model.

\subsubsection{Higher Order CBFs} It is worth mentioning that for the acceleration-controlled nonholonomic models \eqref{eqn:Acceleration controlled Unicycle model}, we can use 
another class of CBFs introduced specifically for constraints with higher relative degrees: HOCBF \cite{7524935,9029455,9456981} given by:  
\begin{align}
h_2 = \dot h_1 + \kappa(h_1),
\label{eq:hocbf}
\end{align}
where $h_1$ is the equation of ellipse given by \eqref{eqn:Ellipse-CBF}. 
\par Apart from lacking geometrical intuition, $h_2$ for acceleration-controlled unicycle model will result in a conservative, safe set as per \cite[Theorem 3]{9516971}. For acceleration controlled bicycle model \eqref{eq:bicyclemodel} with same HOCBF $h_2$ \eqref{eq:hocbf}, if $\mathcal{L}_{g} h_2 = 0$ then we can choose $\dot c_x , \dot c_y$ in such a way that $\mathcal{L}_f h_2 + \frac{\partial h_2}{\partial t} \ngeq 0$ which results in an invalid CBF. Due to space constraints, a detailed proof of the same is omitted and will be explained as part of future work. The results are summarised in Table-\ref{table: types of CBFs}. 

\par However, our goal in this paper is to develop a CBF formulation with geometrical intuition that provides safety guarantees to avoid moving obstacles with the acceleration-controlled nonholonomic models. We propose this next.

\section{ Collision Cone CBF}
\label{section: Collision Cone CBF}

After highlighting the limitations of current collision avoidance methods, we will now introduce our proposed approach, which we refer to as Collision Cone Control Barrier Functions (C3BFs).

A collision cone, defined for a pair of objects, represents a predictive set used to assess the potential for a collision between them based on their relative velocity direction. Specifically, it signifies the directions in which either object, if followed, would lead to a collision between the two.

Throughout this paper, we consider obstacles as ellipses, treating the vehicle as a point. Henceforth, when we mention the term "collision cone", we are referring to this scenario, with the center of the Unmanned Ground Vehicle (UGV) as the point of reference.

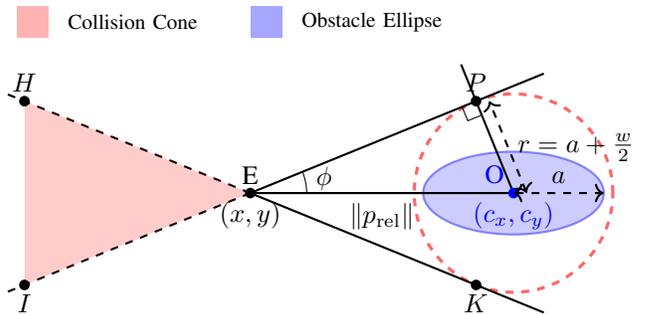
\begin{figure}
    \centering
    \begin{tikzpicture}[
      collisioncone/.style={shape=rectangle, fill=red, line width=2, opacity=0.30},
      obstacleellipse/.style={shape=rectangle, fill=blue, line width=2, opacity=0.35},
    ]
        
        \def\r{1.32003}; 
        \def\q{-3.5}; 
        \def\x{{\r^2/\q}}; 
        \def\y{{\r*sqrt(1-(\r/\q)^2}}; 
        \def\z{{\q - abs(\q - (\r^2/\q))}};
        \coordinate (Q) at (\q,0); 
        \coordinate (P) at (\x,\y); 
        \coordinate (O) at (0.0, 0); 
        \coordinate (E) at (\q, 0); 
        \coordinate (K) at (\x, {-\y}); 
        \coordinate (H) at (\z, \y);
        \coordinate (I) at (\z, {-\y});
        
        \draw[name path = aux, red!60, very thick, dashed] (O) circle (1.32003);
        \draw[blue!50, thick, fill=blue!20] (O) ellipse (1.20 and 0.55);
        \draw[black, thick] (E) -- (O) node [midway, below] {$\|\prel\|$};
        
        
        \draw[black, thick, name path = tangent] ($(Q)!-0.0!(P)$) -- ($(Q)!1.3!(P)$); 
        \draw[black, thick, name path = normal] ($(O)!-0.0!(P)$) -- ($(O)!1.4!(P)$);
        \draw[black, thick] ($(Q)!-0.0!(K)$) -- ($(Q)!1.3!(K)$);
        
        \draw[black, thick, name path = tangent, dashed] ($(Q)!-0.0!(H)$) -- ($(Q)!1.1!(H)$);
        \draw[black, thick, dashed] ($(Q)!-0.0!(I)$) -- ($(Q)!1.1!(I)$);
        
        
        \tkzMarkRightAngle[draw=black,size=.2](O,P,Q);
        \tkzMarkAngle[draw=black, size=0.75](O,Q,P);
        \tkzLabelAngle[dist=1.0](O,Q,P){$\phi$};
        
        \path[shade, left color=red, right color = red, opacity=0.2] (E) -- (H) -- (I) -- cycle;
        
        \fill [black] (E) circle (2pt) node[anchor=north, black] (n1) {$(x,y)$} node[anchor=south, black] (n1) {E};
        \fill [blue] (O) circle (2pt) node[anchor=north, blue] (n1) {$(c_x, c_y)$} node[anchor=south east, blue] (n1) {O}; 
        \fill [black] (P) circle (2pt) node[anchor=south, black] (n1) {$P$};
        \fill [black] (K) circle (2pt) node[anchor=north, black] (n1) {$K$};
        \fill [black] (H) circle (2pt) node[anchor=south, black] (n2) {$H$};
        \fill [black] (I) circle (2pt) node[anchor=north, black] (n1) {$I$};
        
        \draw [<->, color=black, thick, dashed] ([xshift=5 pt, yshift=0 pt]O) -- ([xshift=5 pt, yshift=0 pt]P) node [midway, right] {$r = a+\frac{w}{2}$};
        \draw [<->, color=black, thick, dashed] (O) -- (1.20, 0) node [midway, above] {$a$};
        
        \matrix [above right, nodes in empty cells, matrix of nodes, column sep=0.5cm, inner sep=6pt] at (current bounding box.north west) {
          \node [collisioncone,label=right:{\footnotesize Collision Cone}] {}; &
          \node [obstacleellipse,label=right:{\footnotesize Obstacle Ellipse}] {}; \\
        };
    \end{tikzpicture}
    \caption{Construction of collision cone for an elliptical obstacle considering the UGV's dimensions (width: $w$).} 
    \label{Fig:Construction of Collision Cone}
\end{figure}

Consider a UGV defined by the system (\ref{eqn: affine control system}) and a moving obstacle (e.g., a pedestrian or another vehicle). This is visually depicted in Figure \ref{Fig:Construction of Collision Cone}. We establish the velocity and positions of the obstacle relative to the UGV. To account for the UGV's dimensions, we approximate the obstacle as an ellipse and draw two tangents from the vehicle's center to a conservative circle that encompasses the ellipse (taking into consideration the UGV's dimensions as $r = \max(c_1, c_2) + \frac{w}{2}$), where $c_1, c_2$ are the dimensions of the ellipse and $w$ is the maximum dimension of the UGV. By over-approximating the obstacle as a circle, we enhance the safety threshold and remove complexity related to the orientation of the obstacle.

For a collision to occur, the relative velocity of the obstacle must be directed towards the UGV. Consequently, the relative velocity vector should not point into the shaded pink region labeled EHI in Figure \ref{Fig:Construction of Collision Cone}, which takes the form of a cone.

Let $\mathcal{C}$ denote this set of safe directions for the relative velocity vector. If there exists a function $h:\mathcal{D}\subseteq \mathbb{R}^n \rightarrow \mathbb{R}$ that satisfies the conditions outlined in {Definition \ref{definition: CBF definition}} on $\mathcal{C}$, then we can assert that a Lipschitz continuous control law derived from the resulting Quadratic Program (QP) (\ref{eqn: CBF QP}) for the system ensures that the vehicle will avoid colliding with the obstacle, even if the reference control signal $u_{ref}$ attempts to steer them towards a collision course.

This innovative approach, which avoids the pink cone region, gives rise to what we term as \textit{Collision Cone Control Barrier Functions (C3BFs)}. 
\subsection{Application to systems}
\subsubsection{Acceleration controlled unicycle model}
\label{section: acc controlled unicycle model cccbf}
We first obtain the relative position vector between the body center of the unicycle and the center of the obstacle. 
Therefore, we have
\begin{align}\label{eq:positionvectorunicycle}
    \prel := \begin{bmatrix}
        c_x - (x_p + l \cos\theta) \\
        c_y - (y_p + l \sin\theta)
    \end{bmatrix}.
\end{align}
Here, $l$ is the distance of the body center from the differential drive axis (see Fig. \ref{fig:models}). We obtain its velocity as
\begin{align}\label{eq:velocityvectorunicycle}
    \vrel := \begin{bmatrix}
        \dot c_x - (v \cos \theta - l \sin\theta*\omega) \\
        \dot c_y - (v \sin \theta + l \cos\theta*\omega)
    \end{bmatrix}.
\end{align}
We propose the following CBF candidate:
%
\begin{equation}
    h(\state, t) = <\prel,\vrel> + \|\prel\|\|\vrel\|\cos\phi,
    \label{eqn:CC-CBF}
\end{equation}
where $\phi$ is the half angle of the cone, the expression of $\cos\phi$ is given by $\frac{\sqrt{\|\prel\|^2 - r^2}}{\|\prel\|}$ (see Fig. \ref{Fig:Construction of Collision Cone}). $<.\:,\:.>$ denotes the inner product of two vectors.
The constraint simply ensures that the angle between $\prel, \vrel$ is less than $180^\circ - \phi$.  
We have the following first result of the paper:
%
\begin{theorem}\label{thm:unicycletheorem}{\it
Given the acceleration controlled unicycle model \eqref{eqn:Acceleration controlled Unicycle model}, the proposed CBF candidate \eqref{eqn:CC-CBF} with $\prel,\vrel$ defined by \eqref{eq:positionvectorunicycle}, \eqref{eq:velocityvectorunicycle} is a valid CBF defined for the set $\mathcal{D}$.
}
\end{theorem}
Please refer to \cite[Thm 1]{thontepu2023control} for the proof of Theorem \ref{thm:unicycletheorem}
\subsubsection{Acceleration controlled Bicycle model}
\label{section: bicycle model cccbf}
For the approximated bicycle model \eqref{eqn:bicyclemodel with small beta}, we define the following:
\begin{align}\label{eq:positionvectorbicycle}
    \prel := &
    \begin{bmatrix}
        c_x - x_p &       c_y - y_p 
    \end{bmatrix}^T \\
\label{eq:velocityvectorbicycle}
 \vrel := & \begin{bmatrix}
        \dot c_x - v \cos \theta &        \dot c_y - v \sin \theta 
    \end{bmatrix}^T.
\end{align}
Here, $\vrel$ is NOT equal to the relative velocity $\preldot$. However, for small $\beta$, we can assume that $\vrel$ is the difference between obstacle velocity and the velocity component along the length of the vehicle $v \cos \beta \approx v$. In other words, the goal is to ensure that this approximated velocity $\vrel$ is pointing away from the cone. This is an acceptable approximation as $\beta$ is small and the obstacle radius chosen was conservative (see Fig. \ref{Fig:Construction of Collision Cone}). 
We have the following result:
\begin{theorem}\label{thm:bicycletheorem}{\it
Given the bicycle model \eqref{eqn:bicyclemodel with small beta}, the proposed candidate CBF \eqref{eqn:CC-CBF} with $\prel,\vrel$ defined by \eqref{eq:positionvectorbicycle}, \eqref{eq:velocityvectorbicycle} is a valid CBF defined for the set $\mathcal{C}$.}
\end{theorem}
Please refer to \cite[Thm 2]{thontepu2023control} for the proof of Theorem \ref{thm:bicycletheorem}

\subsubsection{Point mass model}
\label{section: point model cccbf}
\par Similar to Theorem \ref{thm:bicycletheorem}, we can establish similar results for simple point mass models \cite{https://doi.org/10.48550/arxiv.2106.05341} of the form:
\begin{equation}
	\begin{bmatrix}
		\dot{p} \\
		\dot{v}
	\end{bmatrix}
	=
        \begin{bmatrix}
            0_{2x2} & I_{2x2} \\
            0_{2x2} & 0_{2x2}
        \end{bmatrix}
        \begin{bmatrix}
            p\\
            v\\
        \end{bmatrix}
	+
	\begin{bmatrix}
            0_{2x2} \\
            I_{2x2} 
	\end{bmatrix}
		u,
    \label{eqn:Acceleration controlled Point mass model}
\end{equation}
where $p$ = [$x_p$, $y_p$]$^T$, $v$ = [$v_x$, $v_y$]$^T$, and $u$ = [$a_x$, $a_y$]$^T$ $\in\mathbb{R}^2$ denotes the position, velocity and acceleration inputs, respectively. The proposed C3BF-QP is indeed a valid CBF, and its proof is straightforward.

\section{Results}
\label{section: Results and Discussions}
\par In this section, we provide the simulation and experimental results to validate the proposed C3BF-QP. We will first demonstrate the results graphically in Python simulations and then on the Copernicus UGV (modeled as a unicycle) and the FOCAS-Car (modeled as a bicycle).

\subsection{Acceleration Controlled Unicycle Model}

In both simulation and experiments, we have considered the reference control inputs as a simple proportional controller formulated as follows:
\begin{equation}
\begin{aligned}
\label{eqn: Reference controls}
u_{ref}(x) = \begin{bmatrix} a_{ref}(x) \\ \alpha_{ref} (x)
\end{bmatrix}  = \begin{bmatrix} k_{1}*(v_{des} - v) \\
	- k_{2}*\omega \end{bmatrix},
\end{aligned}
\end{equation}
where $k_1,k_2$ are constant gains, $v_{des} \leq v_{max}$ is the target velocity of vehicle. We chose constant target velocities for verifying the C3BF-QP. For the class $\mathcal{K}$ function in the CBF inequality, we chose $\kappa(h) = \gamma h$, where $\gamma=1$. 

Note that the reference controller can be replaced by any existing trajectory tracking / path-planning / obstacle-avoiding controller like the Stanley controller \cite{4282788} or MPC \cite{9483029}. Thus, by integrating the C3BF safety filter with existing state-of-the-art algorithms, we can provide formal guarantees on their safety concerning obstacle avoidance. 

A virtual perception boundary was incorporated, considering the maximum range for the perception sensors. As soon as an obstacle is detected within the perception boundary, the C3BF-QP is activated with $u_{ref}$ given by \eqref{eqn: Reference controls}. The QP yields the optimal accelerations, which are then applied to the robot. 

\subsubsection{Python Simulations}
\label{section: Python Simulation Unicycle model}
\begin{figure}
    \centering
    \includegraphics[width=1\linewidth]{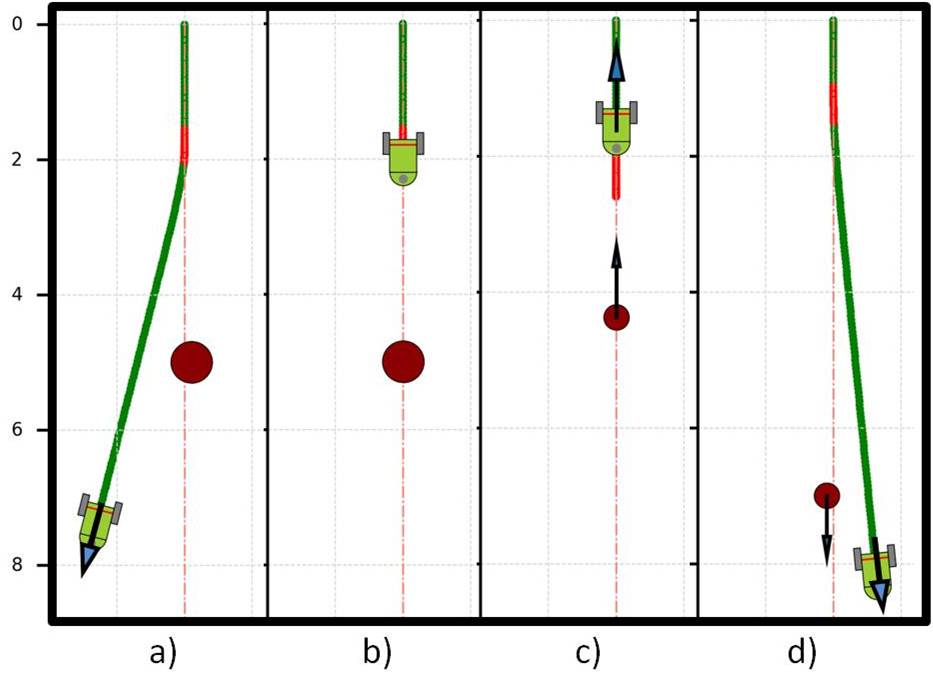}
\caption{Different acceleration-controlled unicycle model behaviors with static (a,b) \& moving (c,d) obstacle. a) turning; b) braking; c) reversing; d) overtaking. The orange lines represent the points where C3BF is active.}
\label{fig: Python Unicycle Behaviours}
\end{figure}

\begin{figure}
    \centering
    \includegraphics[width=\linewidth]{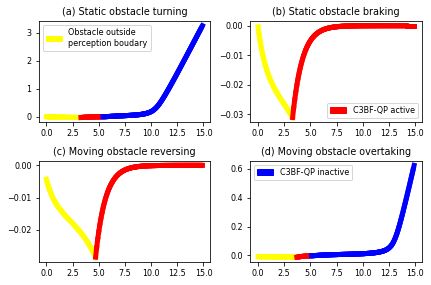}
    \caption{CBF value (h) (Y-axis) varying with time (X-axis). 
    C3BF-QP active means $u_{safe} \neq 0$ (Red) and
    C3BF-QP inactive means $u_{safe} = 0$ (Blue).}
\label{fig: h_graphs of python simulations}
\end{figure}

We consider different scenarios with different poses and velocities of both the vehicle and the obstacle. Different scenarios include static obstacles resulting in a) turning, b) braking and moving obstacles resulting in c) reversing, and d) overtaking, as shown in Fig. \ref{fig: Python Unicycle Behaviours}.
%
The corresponding evolution of CBF value ($h$) as a function of time is shown in Fig. \ref{fig: h_graphs of python simulations}. We can observe that even if $h < 0$ at t = 0 (Fig. \ref{fig: h_graphs of python simulations}(b), (c)), the magnitude is exponentially decreasing and becoming positive.

\subsubsection{Hardware Experiments}
\begin{figure}
    \centering
    \includegraphics[width=1\linewidth]{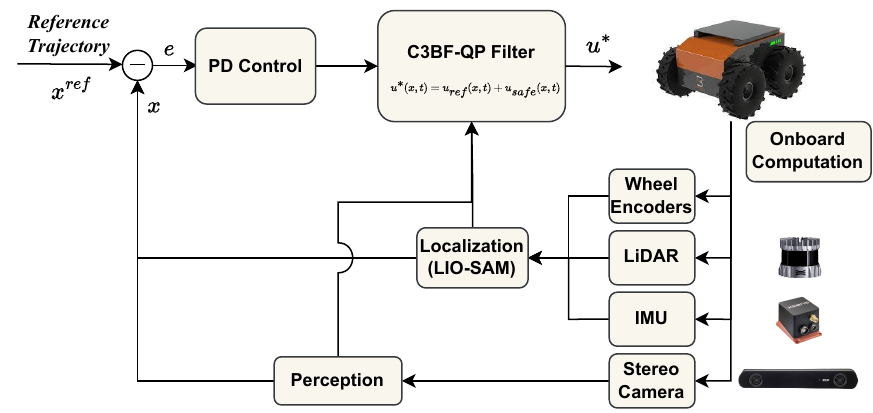}
\caption{Experimental Setup for Copernicus UGV} 
\label{fig: Experimental_setup_unicycle}
\end{figure}

Corresponding to the cases tested in Python simulation, experiments were performed on Botsync's Copernicus UGV (Fig. \ref{fig:Copernicus and FOCAS-Car}). It is a differential drive (4X4) robot with 860 x 760 x 590 mm dimensions. Apart from its internal wheel encoders, external sensors like Ouster OS1-128 Lidar and Xsens IMU MTI-670G sensor were used for localization and minimizing the odometric error through LIO-SAM algorithm \cite{liosam2020shan}. 
Zed-2 Stereo camera was used to detect the obstacles through masking and image segmentation techniques and to get their positions using depth information as shown in Fig \ref{fig: Experimental_setup_unicycle}.  

Our C3BF algorithm combined with the perception stack was first simulated in the Gazebo environment and then tested on the actual robot. We observed braking, turning, reversal, and overtaking behaviors depending on the initial poses of Copernicus, different positions, and velocities of obstacles. These behaviors were similar to the results obtained from Python simulations. We also experimented with various scenarios with multiple stationary obstacles. Even though the behaviors are sensitive to the initial conditions of the robot and obstacle, collision is avoided in all cases. All the results are shown in the attached video link$^{\ref{note: Videos link}}$.

\begin{figure}[htbp]
    \centering
    \includegraphics[width=1\linewidth]{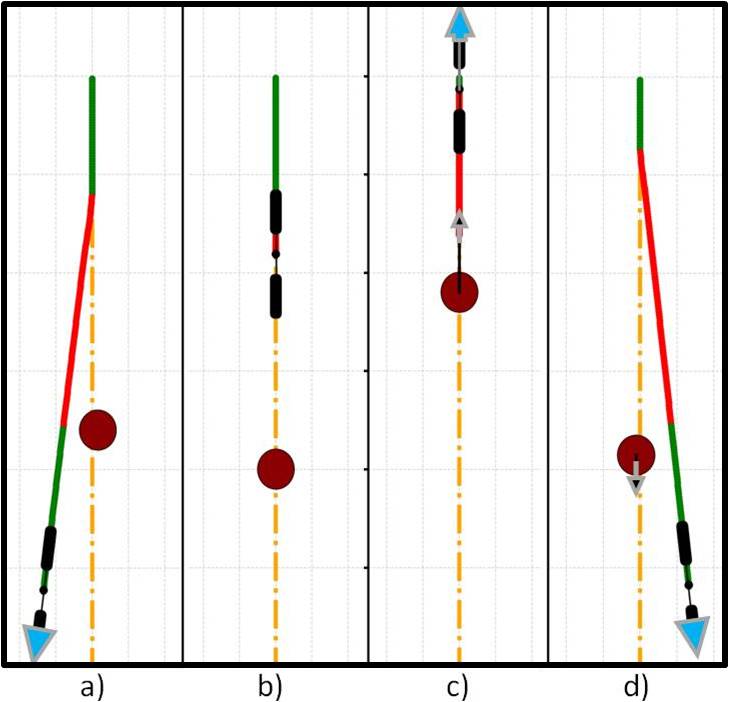}
\caption{Different acceleration-controlled bicycle model behaviors with static (a,b) \& moving (c,d) obstacle. a) turning; b) braking; c) reversing; d) overtaking.} 
\label{fig: Python Bicycle Behaviours}
\end{figure}
\subsection{Acceleration Controlled Bicycle Model}

We have extended and validated our C3BF algorithm for the bicycle model \eqref{eqn:bicyclemodel with small beta} which is a good approximation of actual car dynamics in low-speed scenarios where the lateral acceleration is small ($\leq 0.5\mu g$, $\mu$ is the friction co-efficient) \cite{https://doi.org/10.48550/arxiv.2103.12382} \cite{7995816}. We first simulated all the same four cases without any trajectory tracking reference controller as we did for the unicycle model (Section \ref{section: Python Simulation Unicycle model}) and got similar results as shown in Fig. \ref{fig: Python Bicycle Behaviours}. For the second set of simulations, the reference acceleration $a_{ref}$ was obtained from a P-controller tracking the desired velocity, while the reference steering $\beta_{ref}$ was obtained from the Stanley controller \cite{4282788}. The Stanley controller is fed an explicit reference spline trajectory for tracking, with an obstacle on it. The reference controllers were integrated with the C3BF-QP and applied to the robot simulated in Python (Fig. \ref{fig:python_sim2}) and CARLA simulator.

\subsubsection{Python Simulations}

\par
\begin{figure}[htbp]
    \centering
    \begin{subfigure}[b]{0.48\textwidth}
        \centering
        \begin{tikzpicture}[
          collisioncone/.style={shape=rectangle, fill=red, line width=1, opacity=0.20},
          obstacleellipse/.style={shape=rectangle, fill=green, line width=1, opacity=0.35},
          egovehicle/.style={shape=cross, green, thin}
        ]
        \node (grid_img) {
        \setlength{\tabcolsep}{1pt}
        \begin{tabular}{cc}
             \includegraphics[width=1.6in]{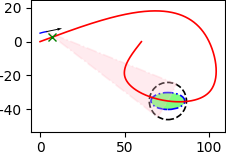}&
             \includegraphics[width=1.6in]{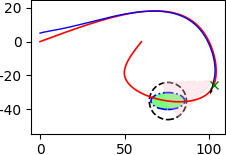}\\
             \includegraphics[width=1.6in]{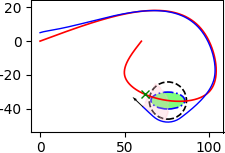}&
             \includegraphics[width=1.6in]{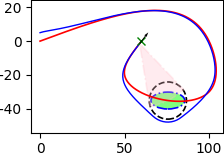}
        \end{tabular}
        };
        \node (grid_legend) at ([xshift=55 pt, yshift=32 pt]grid_img.south west) {\includegraphics[width=0.8in]{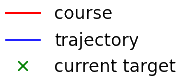}};
        
        \matrix [above, nodes in empty cells, matrix of nodes, column sep=0.3cm, inner sep=2pt] at ([xshift=48 pt, yshift=15 pt]grid_img.south) {
          \node [obstacleellipse,label=right:{\scriptsize \texttt Obstacle}] {}; \\
          \node [collisioncone,label=right:{\scriptsize \texttt Collision Cone}] {}; \\
        };
        \end{tikzpicture}
        \vspace{-7mm}
        \caption{Ego Vehicle avoiding static obstacle (black arrow: $-\vrel$).}
    \end{subfigure}
    \begin{subfigure}[b]{0.4\textwidth}
        \centering
        \includegraphics[width = \linewidth]{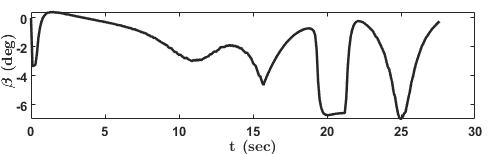}
        \caption{Variation of slip angle ($\beta$) w.r.t. time.}
        \label{fig:python_sim2 beta plot}
    \end{subfigure}
    \caption{Graphical illustration of C3BF on bicycle model (\ref{eqn:bicyclemodel with small beta}). 
    }
    \label{fig:python_sim2}
\end{figure}

 Fig. \ref{fig: Python Bicycle Behaviours} shows different behaviors (turning, braking, reversing, overtaking) of the vehicle modeled as a bicycle model without reference trajectory tracking. Fig. \ref{fig:python_sim2} shows the trajectory setup along with an obstacle placed towards the end. The vehicle tracks the reference spline trajectory (red), and the reflection of the collision cone is shown by the pink region. The C3BF algorithm preemptively prevents the relative velocity vector from falling into the unsafe set, thus avoiding obstacles by circumnavigating. It can be verified from Fig. (\ref{fig:python_sim2 beta plot}) that the slip angle $\beta$ remains small, which is in line with the assumption used \eqref{eqn:bicyclemodel with small beta}. 

\subsubsection{CARLA Simulations}
\begin{figure}
    \centering
    \includegraphics[width=1\linewidth]{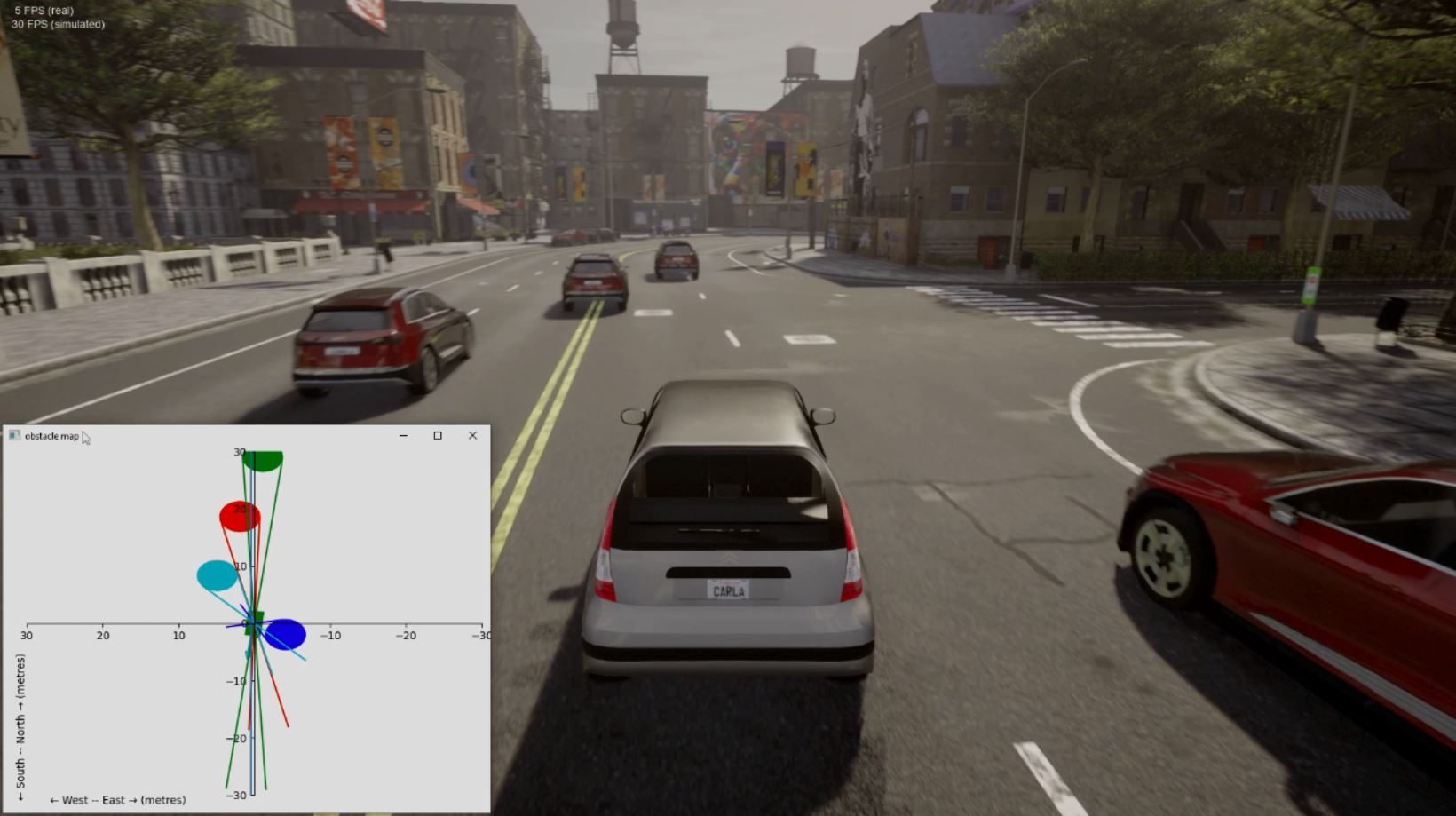}
\caption{Ego-vehicle (white) avoiding multiple moving cars (Red) using C3BF controller with graphical collision cones.}
\label{fig: Carla Simulation}
\end{figure}

\par With the small $\beta$ approximation and using the formulation from Section \ref{section: bicycle model cccbf}, numerical simulations (using \textit{CVXOPT} library) were performed on virtual car model on CARLA simulator \cite{Dosovitskiy17} to demonstrate the efficacy of C3BF in collision avoidance in real-world like environments (Fig. \ref{fig: Carla Simulation}).
A virtual perception boundary was used to accommodate the range of perception sensors on real self-driving cars.
Information about the obstacle is programmatically retrieved from the CARLA server. 
The resulting simulation experiments can be viewed in the attached video link$^{\ref{note: Videos link}}$.

\subsubsection{Hardware Experiments}
\begin{figure}
    \centering
    \includegraphics[width=0.9\linewidth]{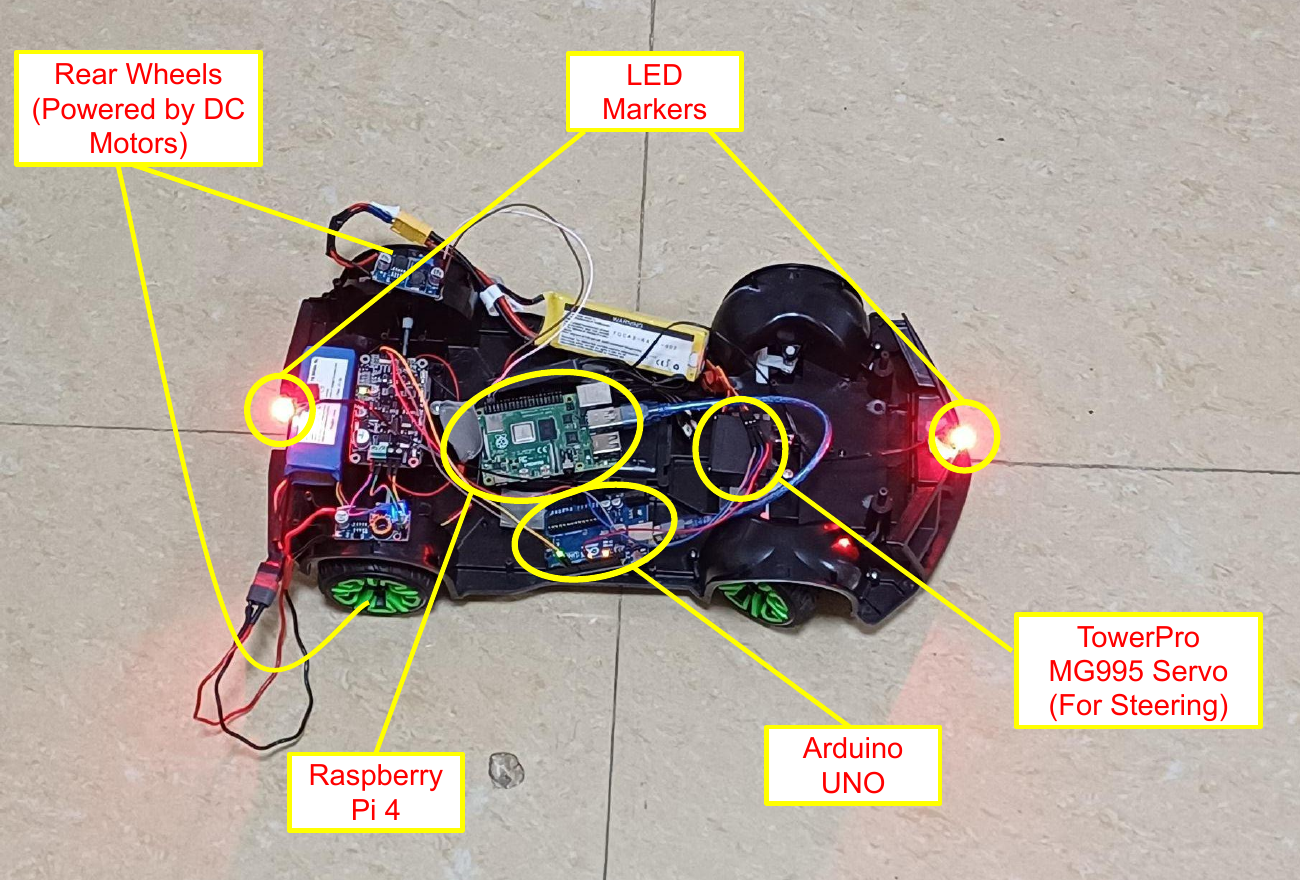}
\caption{FOCAS-Car Hardware Setup} 
\label{fig: Focas-car_hardware}
\end{figure}

Corresponding to the cases tested in Python simulation, experiments were performed on FOCAS-Car (Fig. \ref{fig:Copernicus and FOCAS-Car}), a rear wheel drive car $1/10th$ scaled model, which is powered by 11.1 \textit{V}, 1800 \textit{mAh} LIPO Battery. The steering actuation is handled by a servo motor TowerPro MG995, and a DC motor actuates the rear wheels.
The realization of the proposed controller has been divided into two stages: a high-level controller running ROS (Robotic Operating System) on Ubuntu and a low-level controller realized by an Arduino UNO board. The sampling rate for the ROS master is set to 100 Hz. The Raspberry Pi 4 board runs Ubuntu 20.04 LTS and ROS Noetic on the processor. It is mounted on the car and solves the optimization problem (C3BF-QP) online. The controller has been coded as a ROS Node in Python. At the low level, this controller has less computation, and it runs at a frequency of 57600 BaudRate. 
The global position of the car, as well as the obstacle(s), is measured using PhaseSpace\textsuperscript{\texttrademark} motion capture system with a tracking frequency of 960 $Hz$. The 2 LED Markers are placed in front and rear of the car as shown in Fig. \ref{fig: Focas-car_hardware} to estimate the state of the car ($p, v, \theta, \omega$).

Similar to the experiments conducted in the Python case, we observed braking, turning, and overtaking behaviors. These behaviors were identical to the results obtained from Python simulations. Irrespective of the initial conditions of the robot and the obstacle, collision is avoided in all cases.

\begin{figure}
    \centering
    \includegraphics[width=1\linewidth]{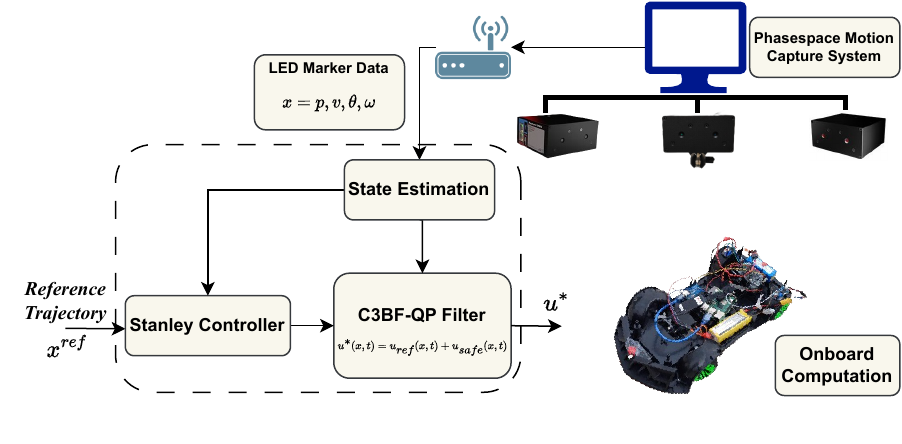}
\caption{Experimental Setup for FOCAS-Car.} 
\label{fig: Experimental_setup_bicycle}
\end{figure}

All the simulations and hardware experiments can be viewed in the attached video link\footnote{\label{note: Videos link}.\url{https://sites.google.com/iisc.ac.in/c3bf-ugv}}

\section{Conclusions}
\label{section: Conclusions}
In this paper, we presented a novel real-time control methodology for Unmanned Ground Vehicles (UGVs) for avoiding moving obstacles by using the concept of collision cones. This idea of constructing a cone around an obstacle is a well-known technique used in planning \cite{Fiorini1993,doi:10.1177/027836499801700706,709600, Chakravarthy2012}, and this enables the vehicle to check the existence of any collision with moving objects. We have extended the same idea with control barrier functions (CBFs) with acceleration-controlled nonholonomic UGV models (unicycle and bicycle models), thereby enabling fast modification of reference control inputs giving guarantees of collision avoidance in real-time.
The proposed QP formulation (C3BF-QP) acts as a filter on the reference controller allowing the vehicle to safely maneuver under different scenarios presented in the paper. Our C3BF safety filter can be integrated with any existing state-of-the-art trajectory tracking/path-planning/obstacle-avoiding algorithms to ensure real-time safety by giving formal guarantees. We validated the proposed C3BF controller in simulations as well as hardware by integrating it with simple trajectory-tracking reference controllers. 
As a part of future work, the focus will be more on self-driving vehicles with complex dynamic models for avoiding moving obstacles on the road. 


\label{section: References}
\bibliographystyle{IEEEtran}
\bibliography{references.bib}

\end{document}